# IMAGE DYNAMIC RANGE ENHANCEMENT IN THE CONTEXT OF LOGARITHMIC MODELS

Vasile Patrascu, Vasile Buzuloiu

University " Politehnica" Bucureşti

E-mail: vpatrascu@tarom.ro, buzuloiu@alpha.imag.pub.ro

## ABSTRACT

Images of a scene observed under a variable illumination or with a variable optical aperture are not identical. Does a privileged representant exist? In which mathematical context? How to obtain it? The authors answer to such questions in the context of logarithmic models for images. After a short presentation of the model, the paper presents two image transforms: one performs an optimal enhancement of the dynamic range, and the other does the same for the mean dynamic range. Experimental results are shown.

## 1 INTRODUCTION

Image enhancement is an important branch of image processing. Many approaches exist (contrast manipulation, histogram modification, filtering,..) that are well exposed in reference books such as [1-3, 10, 11]. Stockham [12] proposed an image enhancement method based on the homomorphic theory introduced by Oppenheim [6] and applied to images obtained by transmitted or reflected light. The key of this approach is to use an adapted mathematical homomorphism, that performs a transformation in order to use the classical linear mathematics and to use linear image processing techniques. Another approach exists in the general setting of logarithmic representation suited for the transmitted light imaging processes or the human visual perception. Jourlin and Pinoli introduced a mathematical framework for this kind of "non-linear" representations [4, 5, 9]. In this paper we present another slightly different logarithmic model which permits to maximize the dynamic range. Also we present new formulas for mean dynamic range. The remainder of the paper is organized as follows: Section 2 introduces the addition, the real scalar multiplication, and the product of two gray levels. Similarly, the Section 3 introduces the addition, the real scalar multiplication, and the product for the gray level images. Section 4 defines two optimal image transforms using our mathematical model. Section 5 presents experimental results and Section 6 outlines the conclusions.

## 2 THE REAL ALGEBRA OF THE GRAY LEVELS

We consider as the space of gray levels, the set $E = (0, \infty)$. Let be $M$ a real and positive number. In the set of gray levels $E$ we will define the addition $\langle + \rangle$ and the real scalar multiplication $\langle \times \rangle$.

### 2.1 Addition

$\forall v_1, v_2 \in E$ the sum $v_1 \langle + \rangle v_2$ is done by the following relation:
$$v_1 \langle + \rangle v_2 = \frac{v_1 \cdot v_2}{M} \quad (1)$$

The neutral element for addition is $\theta = M$.

Each element $v \in E$ has as its opposite the element $w = \frac{M^2}{v}$ and this verifies the following equation: $v \langle + \rangle w = \theta$.

The addition $\langle + \rangle$ is stable, associative, commutative, has a neutral element and each element has an opposite. It results that this operation establishes on $E$ a commutative group structure.

We can also define the subtraction operation $\langle - \rangle$ by:
$$v_1 \langle - \rangle v_2 = \frac{v_1 \cdot M}{v_2} \quad (2)$$

Using subtraction $\langle - \rangle$, we will note the opposite of $v$, with $\langle - \rangle v$.

### 2.2 Scalar multiplication

For $\forall \lambda \in R, \forall v \in E$, we define the product between $\lambda$ and $v$ by:
$$\lambda \langle \times \rangle v = M \cdot \left( \frac{v}{M} \right)^{\lambda} \quad (3)$$

The two operations: addition $\langle + \rangle$ and scalar multiplication $\langle \times \rangle$ establish on $E$ a real vector space structure.

### 2.3. The product operation

For $\forall v_1, v_2 \in E$ the product $v_1 \langle \cdot \rangle v_2$ is defined by the relation:
$$v_1 \langle \cdot \rangle v_2 = M \cdot e^{M \cdot \ln\left( \frac{v_1}{M} \right) \cdot \ln\left( \frac{v_2}{M} \right)} \quad (4)$$

The neutral element for product is $u = M \cdot e^{\frac{1}{M}}$.

The three operations, addition $\langle+\rangle$, scalar multiplication $\langle\times\rangle$ and product $\langle\cdot\rangle$ establish on E a real algebra structure.

## 3 THE REAL ALGEBRA OF THE GRAY LEVEL IMAGES

A gray level image is a function defined on a bi-dimensional compact D from $R^2$ taking the values in the gray level space E. We note with F(D,E) the set of gray level images defined on D. We can extend the operations defined on E to gray level images F (D,E), in a natural way:

### 3.1 Addition
$\forall f_1, f_2 \in F(D,E), \forall (x,y) \in D$,
$$(f_1 \langle+\rangle f_2)(x,y) = f_1(x,y) \langle+\rangle f_2(x,y) \quad (5)$$

The neutral element is the function $f(x,y) = M$ for $\forall (x,y) \in D$. The addition $\langle+\rangle$ is stable, associative, commutative, has a neutral element and each element has an opposite. As a conclusion, this operation establishes on the set F(D,E) a commutative group structure.

### 3.2. Scalar multiplication
$\forall \lambda \in R, \forall f \in F(D,E), \forall (x,y) \in D$,
$$(\lambda \langle\times\rangle f)(x,y) = \lambda \langle\times\rangle f(x,y) \quad (6)$$

The two operations, addition $\langle+\rangle$ and scalar multiplication $\langle\times\rangle$ establish on F(D,E) a real vector space structure.

### 3.3. The product operation
For $\forall f_1, f_2 \in F(D,E), \forall x,y \in D$ the product $f_1 \langle\cdot\rangle f_2$ is defined by the relation:
$$(f_1 \langle\cdot\rangle f_2)(x,y) = M \cdot e^{M \cdot \ln\left(\frac{f_1(x,y)}{M}\right) \cdot \ln\left(\frac{f_2(x,y)}{M}\right)} \quad (7)$$

The neutral element for product is the function $u(x,y) = M \cdot e^{\frac{1}{M}}, \forall (x,y) \in D$. The three operations, addition $\langle+\rangle$, scalar multiplication $\langle\times\rangle$ and product $\langle\cdot\rangle$ establish on F(D,E) a real algebra structure.

## 4 GRAY LEVEL IMAGE ENHANCEMENT

### 4.1 Enhancement of the dynamic range
Let f be an image defined on the spatial domain D. Let us denote $f_i$ and $f_s$ the lower bound and the upper bound on D, respectively,
$$f_i = \inf_{x \in D} f(x) \text{ and } f_s = \sup_{x \in D} f(x). \quad (8)$$

The dynamic range of f, denoted $D_t(f)$, is defined [5] as the (real) difference
$$D_t(f) = f_s - f_i \quad (9)$$
where the subtraction is meant in R

A $\lambda$-homothetic of f is defined as $\lambda \langle\times\rangle f$ and consequently the dynamic range of positive homothetic is
$$D_t(\lambda \langle\times\rangle f) = (\lambda \langle\times\rangle f)_s - (\lambda \langle\times\rangle f)_i = \lambda \langle\times\rangle f_s - \lambda \langle\times\rangle f_i \quad (10)$$

where $\lambda$ is a positive real number. The class $(\lambda \langle\times\rangle f)_{\lambda>0}$ of strictly positive homotetics associated with an image f appears naturally as the set of reference where the solution is to be found.

The optimization problem is to find the positive homothetic with the larger dynamic range, supposing that the lower bound and the upper bound of the image satisfy the natural inequalities:
$$0 < f_i < f_s < M. \quad (11)$$

Under the previous conditions, there exists a unique strictly positive real number, denoted $\lambda_t(f)$, called the optimal logarithmic gain, such that the image $\lambda_t(f) \langle\times\rangle f$ presents the maximal dynamic range in the $(\lambda \langle\times\rangle f)_{\lambda>0}$ class, namely
$$D_t(\lambda_t(f) \langle\times\rangle f) = \max_{\lambda>0}\left(D_t(\lambda \langle\times\rangle f)\right) \quad (12)$$

Such a strictly positive real number is explicitly defined by
$$\lambda_t(f) = \frac{\ln\left(\frac{\ln(M/f_i)}{\ln(M/f_s)}\right)}{\ln(f_s/f_i)} \quad (13)$$

So, the image transform $S_t$, that performs the optimal enhancement of the dynamic range, is defined by
$$S_t(f) = \lambda_t(f) \langle\times\rangle f \quad (14)$$

We shall proceed to prove (13):
Define the function $h:(0,\infty) \to [0,M]$ by putting
$$h(\lambda) = \lambda \langle\times\rangle f_s - \lambda \langle\times\rangle f_i, \quad (15)$$
namely,
$$h(\lambda) = M \cdot \left(\frac{f_s}{M}\right)^\lambda - M \cdot \left(\frac{f_i}{M}\right)^\lambda.$$ Solving the equation $h'(\lambda) = 0$ yields a unique solution, $\lambda_t(f)$, such that
$$M \cdot \left(\frac{f_s}{M}\right)^{\lambda_t(f)} \cdot \ln\left(\frac{f_s}{M}\right) - M \cdot \left(\frac{f_i}{M}\right)^{\lambda_t(f)} \cdot \ln\left(\frac{f_i}{M}\right) = 0$$

Finally, one gets equation (13).
The second derivative $h''(\lambda)$ equals
$$h''(\lambda) = M \cdot \left(\frac{f_s}{M}\right)^\lambda \cdot \ln\left(\frac{f_s}{M}\right)^2 - M \cdot \left(\frac{f_i}{M}\right)^\lambda \cdot \ln\left(\frac{f_i}{M}\right)^2 \quad (16)$$

We can observe that $h''(2\lambda_t) = 0$ and $h''(\lambda) < 0$ for $\lambda \in (0, 2\lambda_t)$ and $h''(\lambda) > 0$ for $\lambda \in (2\lambda_t, \infty)$. We can see that in Fig.3 and Fig. 6.

### 4.2 Enhancement of the mean dynamic range
If the dynamic range is near M from the beginning the $S_t$ transform is of no practical use. Nevertheless,

this can happen in practice due to the noise which produces some (few) pixels with 0 value and some (few) others with M value ('salt and pepper" noise). For such cases it is necessary to find a way to eliminate such false values. We propose to use the statistical moments of the image. Namely we shall replace the original image with another one having only two values $v_s(f)$ and $v_i(f)$ so that the first 3 moments are preserved [8]. We can define the mean dynamic range by:

$$D_m(f) = v_s(f) - v_i(f) \qquad (17)$$

The following algebraic system defines our values, $v_s(f)$ and $v_i(f)$:

$$\begin{cases} p_i + p_s = 1 \\ p_i \langle \times \rangle v_i(f) \langle + \rangle p_s \langle \times \rangle v_s(f) = m_1(f) \\ p_i \langle \times \rangle (v_i(f))^2 \langle + \rangle p_s \langle \times \rangle (v_s(f))^2 = m_2(f) \\ p_i \langle \times \rangle (v_i(f))^3 \langle + \rangle p_s \langle \times \rangle (v_s(f))^3 = m_3(f) \end{cases} \qquad (18)$$

where $m_1(f), m_2(f), m_3(f)$ are the statistical moments of order one, two and three for the image f.

We put $\sigma = \sqrt{m_2(f) \langle - \rangle (m_1(f))^2}$ (19)

and

$$\mu = \sqrt[3]{m_3(f) \langle - \rangle 3 \langle \times \rangle m_2(f) \langle \cdot \rangle m_1(f) + 2 \langle \times \rangle (m_1(f))^3}$$
(20)

It results the following values:

$$v_i(f) = m_1(f) \langle + \rangle \frac{1}{2} \langle \times \rangle \sigma^{-2} \langle \cdot \rangle \left( \mu^3 \langle - \rangle \sqrt{4 \langle \times \rangle \sigma^6 \langle + \rangle \mu^6} \right)$$
(21)

$$v_s(f) = m_1(f) \langle + \rangle \frac{1}{2} \langle \times \rangle \sigma^{-2} \langle \cdot \rangle \left( \mu^3 \langle + \rangle \sqrt{4 \langle \times \rangle \sigma^6 \langle + \rangle \mu^6} \right)$$
(22)

We do not detail the formulae for the probabilities $p_i$ and $p_s$, as they are of no use here. The new image transform, denoted $S_m$ maximizing the mean dynamic range of an image f is defined by

$$S_m(f) = \lambda_m(f) \langle \times \rangle f \qquad (23)$$

where $\lambda_m(f)$ is the unique strictly positive real number, called the optimal mean logarithmic gain given by

$$\lambda_m(f) = \frac{\ln\left(\frac{\ln(M/v_i(f))}{\ln(M/v_s(f))}\right)}{\ln(v_s(f)/v_i(f))} \qquad (24)$$

The proof is similarly to the previous one.

## 5. EXPERIMENTAL RESULTS

The images are taken from [7].
Values obtained for image "cells" are:
$v_i = 244.78$, $v_s = 251.91$,
$\lambda_m = 35.66$, $D_m(\lambda_m) = 92.465$.
Values obtained for image "ball" are:

$v_i = 5.886$, $v_s = 80.324$,
$\lambda_m = 0.4515$, $D_m(\lambda_m) = 105.07$.

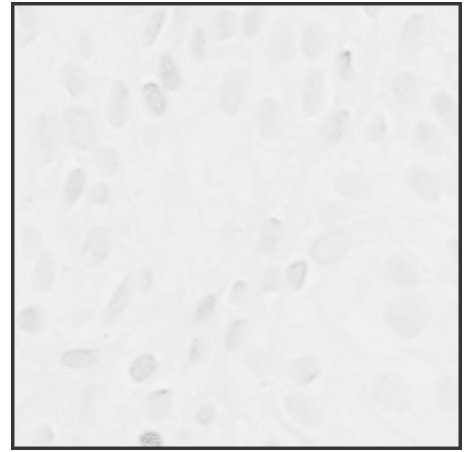

Fig.1 Image "cells"

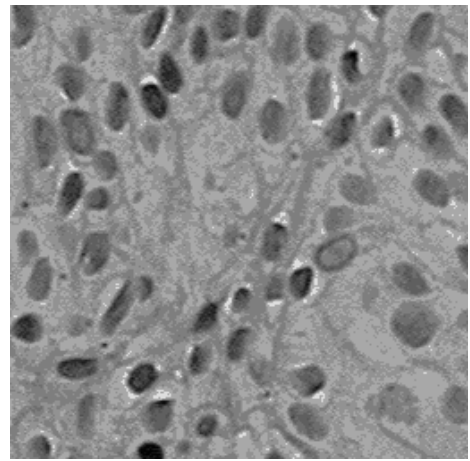

Fig.2 Enhanced image "cells"

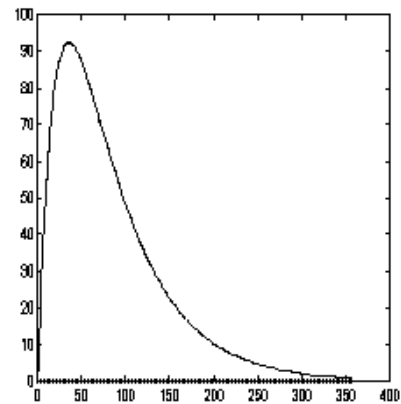

Fig.3 Function $D_m(\lambda)$ for image "cells"

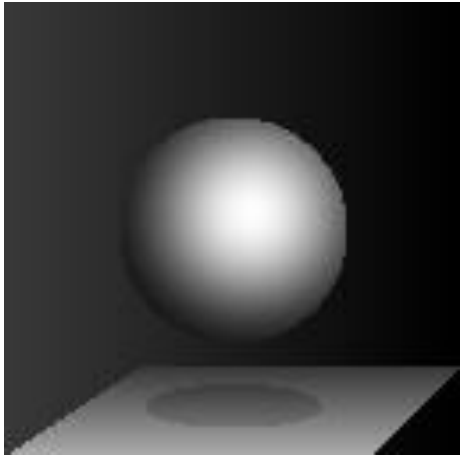

Fig.4   Image "ball"

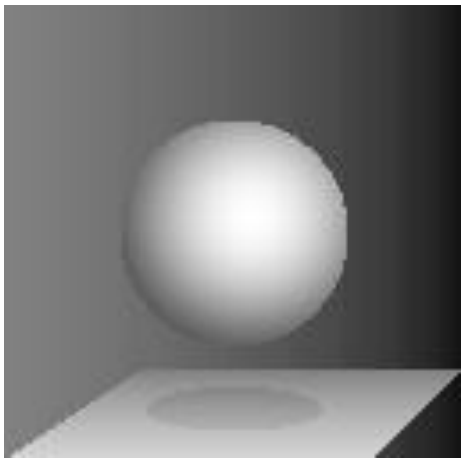

Fig. 5  Enhanced image "ball"

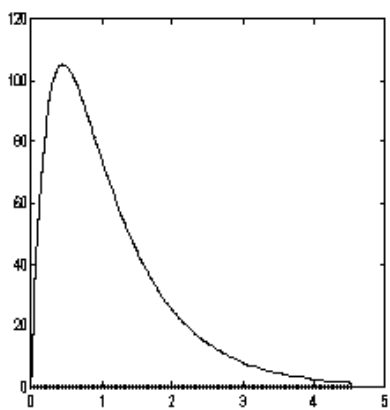

Fig.6 Function $D_m(\lambda)$ for image "ball"

## 6 CONCLUSION

The image enhancement has been and is always an important field of the research where publications of new techniques and methods take a large places in the literature. A great number of enhancement methods exist because each application is specific and needs an adapted method [2]. So, the physical nature of images to be processed is of central importance and the need of an adequate image mathematic model appears clearly as a necessity [5]. In this paper we have presented an image enhancement transformation like gamma correction which needs an adapted mathematical model for becoming optimizable. Also we presented a new formula for the mean dynamic range.